\documentclass{article} % For LaTeX2e
\usepackage{nips13submit_e,times}
\usepackage{hyperref}
\usepackage{url}
\usepackage{graphicx}
\usepackage{color}

\usepackage{url}
\usepackage{amsmath}
\usepackage{amsfonts}
\usepackage{mathrsfs}
\usepackage{amssymb}
\usepackage{amsthm}

\usepackage{subfigure}

\usepackage{algorithm}
\usepackage{algorithmic}
\usepackage{paralist}

\newtheorem{theorem}{Theorem}

\title{Sparse Linear Dynamical System with Its Application in Multivariate Clinical Time Series}

\author{
Zitao Liu\\
Department of Computer Science\\
University of Pittsburgh\\
Pittsburgh, PA 15213 \\
\texttt{ztliu@cs.pitt.edu} \\
\And
Milos Hauskrecht \\
Department of Computer Science\\
University of Pittsburgh\\
Pittsburgh, PA 15213 \\
\texttt{milos@cs.pitt.edu} \\
}

\nipsfinalcopy % Uncomment for camera-ready version

\begin{document}

\maketitle

\vspace{-0.5cm}

\begin{abstract}
Linear Dynamical System (LDS) is an elegant mathematical framework for modeling and learning multivariate time series. However, in general, it is difficult to set the dimension of its hidden state space. A small number of hidden states may not be able to model the complexities of a time series, while a large number of hidden states can lead to overfitting. In this paper, we study methods that impose an $\ell_1$ regularization on the transition matrix of an LDS model to alleviate the problem of choosing the optimal number of hidden states. We incorporate a generalized gradient descent method into the Maximum a Posteriori (MAP) framework and use Expectation Maximization (EM) to iteratively achieve sparsity on the transition matrix of an LDS model. We show that our Sparse Linear Dynamical System (SLDS) improves the predictive performance when compared to ordinary LDS on a multivariate clinical time series dataset.    
\end{abstract}

\vspace{-0.3cm}

\section{Introduction}
\label{sec:intro}

\vspace{-0.2cm}

Developing accurate models of dynamical systems is critical for their successful applications in outcome prediction, decision support, and optimal control. A large spectrum of models have been developed and successfully applied for these purposes in the literature \cite{bataltemporal,hamilton1994time,ljung1994modeling}. In this paper we focus on a popular model for time series analysis: the Linear Dynamical System (LDS) \cite{kalman1960new} and its application to clinical time series \cite{liu2013clinical,liu2013modeling}. We aim to develop a method to learn an LDS that performs better on future value predictions when learned from a small amount of complex multivariate time series dataset. 
\nocite{batal2012mining,batal2011pattern}

LDS is a widely used model for time series analysis of real-valued sequences. The model is Markovian and assumes the dynamic behaviour of the system is captured well using a small set of real-valued hidden-state variables and linear-state transitions corrupted by a Gaussian noise. The observations in LDS, similarly to hidden states, are real-valued. Briefly, the observations at time $t$ are linear combinations of hidden state values for the same time. While in some LDS applications the model parameters are known a priori, in the majority of real-world applications the model parameters are unknown, and we need learn them from data that consists of observation sequences we assume were generated by the LDS model. While this can be done using standard LDS learning approaches, the problem of learning an LDS model gives rise to numerous important questions: Given the multivariate observation sequences, how many hidden states are needed to represent the system dynamics well? Moreover, since transition and observation matrices depend on the number of hidden states, how do we prevent the overfit of the model parameters when the number of examples is small? 

In this work we address the above issues by presenting a method based on the sparse representation of LDS (SLDS) that is able to adjust (depending on the observation sequences in the data) the number of hidden states and at the same time prevent the overfit of the model. Our approach builds upon the probabilistic formulation of the LDS model, and casts the optimization of its parameters as a maximum a posteriori (MAP) estimate, where the choice of the parameter priors biases the model towards sparse solutions. 

Our SLDS approach is distinctly different from previous work \cite{charles2011sparsity, chiuso2010learning, ghanem2010sparse}. \cite{charles2011sparsity} formulates the traditional Kalman filter as a one-step update optimization procedure and incorporates sparsity constraints to achieve sparsity in the hidden states. \cite{ghanem2010sparse} trains an LDS for each training example and tries to find a sparse linear combination of coefficients in order to combine the ensemble of models. Neither \cite{charles2011sparsity} nor \cite{ghanem2010sparse} directly achieve sparsity on the parameters of the LDS, and furthermore, the performance of their resulting models still depends on the optimal number of hidden states. \cite{chiuso2010learning} introduces a Bayesian nonparametric approach to the identification of observation-only linear systems, where no hidden states are involved. The underlining assumption is that the observations are obtained from linear combinations of previous observations and some system inputs, which may be too restrictive to model complex multivariate time series and makes the model more sensitive to noisy observations and outliers.

We test our sparse solution on the problem of modeling the dynamics of sequences of laboratory test results. We show that it improves the learning of the LDS model and leads to better accuracy in predicting future time-series values.

Our paper is organized as follows. In Section \ref{sec:lds} we review the basics of the linear dynamical system. In Section \ref{sec:slds} we describe SLDS -- our method of sparsifying the LDS parameters. Inference and learning details of SLDS are explained in Section 3. Experimental results that compare SLDS method to ordinary LDS are presented in Section \ref{sec:experiments}. In Section \ref{sec:conclusion}, we summarize the work and outline possible future extensions.

\vspace{-0.2cm}

\section{Linear Dynamical System (LDS)}
\label{sec:lds}

\vspace{-0.2cm}

The Linear Dynamical System (LDS) is a real-valued time series model that represents observation sequences indirectly with the help of hidden states. Let $\{{z}_{t}\}$, $\{{y}_{t}\}$ define sequences of hidden states and observations respectively. The LDS models the dynamics of these sequences in terms of the state transition probability $p({z}_t | {z}_{t-1})$, and state-observation  probability $p({y}_t | {z}_t)$. These probabilities are modeled using the following equations: 

\vspace{-0.4cm}

\begin{equation}
	\label{eq:lds}
	{z}_t = A{z}_{t-1} + {e}_t; \hspace{0.5cm}  {y}_t = C{z}_t + {v}_t,
\end{equation}

\vspace{-0.2cm}

\noindent where ${y}_t$ is a $d \times 1$ observation vector made at (current time) $t$, and ${z}_t$ an $l \times 1$ hidden states vector. The transitions among the current and previous hidden states are linear and captured in terms of an $l \times l$ transition matrix $A$. The stochastic component of the transition, ${e}_t$, is modeled by a zero-mean Gaussian noise ${e}_t \sim \mathcal{N}(0, Q)$ with an $l \times 1$ zero mean and an $l \times l$ covariance matrix \emph{Q}. The observations sequence is derived from the hidden states sequence. The dependencies in between the two are linear and modeled using a $d \times l$ emission matrix \emph{C}. A zero mean Gaussian noise ${v}_t \sim \mathcal{N}(0, R)$ models the stochastic relation in between the states and observation. In addition to $A,C,Q,R$, the LDS is defined by the initial state distribution for ${z}_1$ with mean $\boldsymbol{\pi}_1$ and covariance matrix $V_1$, ${z}_1 \sim \mathcal{N}( \boldsymbol{\pi_1}, V_1)$.  The complete set of the LDS parameters is $\Omega = \{A, C, Q, R, \boldsymbol{\pi_1}, V_1\}$. The parameters of the LDS model can be learned using either the Expectation-Maximization (EM) algorithm \cite{ghahramani1996parameter} or spectral learning algorithms \cite{katayama2005subspace,overschee1996subspace}.

\vspace{-0.2cm}

\section{Sparse Linear Dynamical System (SLDS)}
\label{sec:slds}

\vspace{-0.2cm}

In this section, we propose a sparse representation of LDS that is able to adjust the number of hidden states and at the same time prevents the overfit of the model. More specifically, we impose $\ell_1$ regularizers on every element of the transition matrix $A_{ij}$, which leads to zero entries in the transition matrix \emph{A}. The zero entries in the transition matrix of LDS indeed reduce the actual number of parameters of LDS, sparsify the hidden states, and avoid the overfitting problem from the real data, even if we set the number of hidden states originally picked is too large. 

To achieve sparsity on the transition matrix, we introduce a Laplacian prior to each element of \emph{A}, $A_{ij}$, since Laplacian priors are equivalent to $\ell_1$ regularizations \cite{chen1998atomic,guan2009sparse,williams1995bayesian}. In general, the Laplacian distribution has the following form: $p(x|\mu, \lambda) = \frac{1}{2\lambda} \exp ( -\frac{|x-\mu|}{\lambda} )$, $\lambda \geq 0$ where $\mu$ is the location parameter and $\lambda$ is the scale parameter. Here, we assume every element $A_{ij}$ is independent to each other and has the following Laplacian density ($\mu = 0$ and $\lambda = 1/\beta$), $p(A_{ij}|\beta) = \frac{\beta}{2} \exp (-\beta |A_{ij}| ) $. Hence, the prior probability for \emph{A} is $p(A|\beta) = \prod_{i=1}^l \prod_{j=1}^l p(A_{ij}|\beta)$ and the log joint distribution for SLDS is:

\vspace{-0.75cm}

\begin{eqnarray}
    \label{eq:joint}
 \log p(\mathbf{z}, \mathbf{y}, A)  =  \log p(A)  + \log p(z_1) + \sum_{t=2}^T \log p(z_t|z_{t-1}, A) + \sum_{t=1}^T \log p(y_t|z_t)
\end{eqnarray}

\vspace{-0.35cm}

\noindent where \emph{T} is the observation sequence length.

\vspace{-0.2cm}

\subsection{Learning}
\label{sec:learning}

\vspace{-0.2cm}

In this section we develop an EM algorithm for the MAP estimation of the SLDS. Let $\hat{z}_{t|T} \equiv \mathbb{E}[z_t| \mathbf{y}]$, $M_{t|T} \equiv \mathbb{E}[z_tz_t^{'}| \mathbf{y}]$, $M_{t,t-1|T} \equiv \mathbb{E}[z_tz_{t-1}^{'}| \mathbf{y} ]$ and define the $\mathcal{Q}$ function as $\mathcal{Q} = \mathbb{E}_{\mathbf{z}} \Big[\log p(\mathbf{z}, \mathbf{y}, A) | \Omega  \Big] $, where

\vspace{-0.4cm}

\begin{eqnarray}
    \label{eq:Q_func}
    \mathcal{Q} = \mathbb{E}_{\mathbf{z}} \Big[\log p(z_1) \Big]  + \log p(A) + \mathbb{E}_{\mathbf{z}}\Big[\sum_{t=2}^T \log p(z_t|z_{t-1}, A)\Big]  +  \mathbb{E}_{\mathbf{z}}\Big[\sum_{t=1}^T \log p(y_t|z_t)\Big] 
\end{eqnarray}

\vspace{-0.3cm}

In the E-step (Inference), we follow the backward algorithm in \cite{ghahramani1996parameter} to compute $\mathbb{E}[z_t| \mathbf{y}]$, $\mathbb{E}[z_tz_t^{'}| \mathbf{y}]$ and $\mathbb{E}[z_tz_{t-1}^{'}| \mathbf{y}]$, which are sufficient statistics of the expected log likelihood. In the M-step (Learning), we try to find $\Omega$ that maximizes the likelihood lower bound $\mathcal{Q}$. In the following, we derive the M-step for gradient based optimization of the parameters $\Omega$. We omit the explicit conditioning on $\Omega$ for notational brevity. Since the $\mathcal{Q}$ function is non-differentiable with respect to \emph{A}, but differentiable with respect to all the other variables ($C, R, Q,\pi_1, V_1$ ), we separate the optimization into two parts.

\textbf{Optimization of $A$.} In each iteration in the M-step, we need to maximize $\mathbb{E}_{\mathbf{z}}\Big[\sum_{t=2}^T \log p(z_t|z_{t-1}, A)\Big] + \log p(A)$ with respect to \emph{A}, which is equivalent to minimizing a function $f(A)$ that 

\vspace{-0.3cm}

\begin{equation}
    \label{eq:funcA}
    f(A) = \underbrace{\frac12 \sum_{t=2}^T \mathbb{E}_{\mathbf{z}} \Big[
    (z_t-Az_{t-1})'Q^{-1}(z_t-Az_{t-1}) \Big]}_\text{g(A)} + \underbrace{\beta ||A||_1}_\text{h(A)}
\end{equation}

\vspace{-0.2cm}

\noindent where $||A||_1$ is the $\ell_1$ norm on every element of matrix \emph{A}, $||A||_1 = \sum_{i=1}^l \sum_{j=1}^l ||A_{ij}||_1$. 

As we can see $f(A)$ is convex but non-differentiable and we can easily decompose $f(A)$ into two parts: $f(A) = g(A) + h(A)$, as shown in eq.(\ref{eq:funcA}). Since $g(A)$ is differentiable, we can adopt the generalized gradient descent algorithm to minimize $f(A)$. The update rule is: $A^{(k+1)} = \mbox{prox}_{\alpha_k} ( A^{(k)} - \alpha_k  \bigtriangledown g(A^{(k)}))$ where $\alpha_k$ is the step size at iteration \emph{k} and the proximal function $\mbox{prox}_{\alpha_k}(A)$ is defined as the soft-thresholding function $S_{\beta \alpha_k}(A)$

\vspace{-0.3cm}

\begin{equation*}
    [S_{\beta \alpha_k}(A)]_{ij} =  
    \begin{cases}
        A_{ij} - \beta \alpha_k & \text{if } A_{ij} > \beta \alpha_k \\
        0 & \text{if } - \beta \alpha_k \leq A_{ij} \leq \beta \alpha_k \\
        A_{ij} + \beta \alpha_k &  \text{if } A_{ij} < - \beta \alpha_k
    \end{cases}
\end{equation*}

\vspace{-0.3cm}

\begin{theorem}
    \label{thm:lip}
    Generalized gradient descent with a fixed step size $\alpha \leq 1/(||Q^{-1}||_F\cdot ||\sum_{t=2}^T M_{t-1|T}||_F)$ for minimizing eq.(\ref{eq:funcA}) has
    convergence rate $O(1/k)$, where k is the number of iterations.
\end{theorem}

\vspace{-0.3cm}

\begin{proof}
$g(A)$ is differentiable with respect to \emph{A}, and its gradient is $\bigtriangledown g(A) =  Q^{-1} ( A \sum_{t=2}^T M_{t-1|T} - \sum_{t=2}^T M_{t,t-1|T} )$. Using simple algebraic manipulation we arrive at $||\bigtriangledown g(X) - \bigtriangledown g(Y)||_F = ||Q^{-1} (X-Y) \sum_{t=2}^T M_{t-1|T}||_F \leq ||Q^{-1}||_F\cdot||\sum_{t=2}^T M_{t-1|T}||_F\cdot ||X-Y||_F$ where $||\cdot||_F$ is the Frobenius norm and the inequality holds because of the sub-multiplicative property of Frobenius norm. Since we know from eq.(\ref{eq:funcA}), $f(A) = g(A) + \beta ||A||_1$, and $g(A)$ has Lipschitz continuous gradient with constant $||Q^{-1}||_F\cdot ||\sum_{t=2}^T M_{t-1|T}||_F$, according to \cite{fornasier2008iterative, shor1968rate}, $f(A^{(k)}) - f(A^*) \leq ||A^{(0)}-A^*||^2_F/2 \alpha k$, where $A^{(0)}$ is the initial value and $A^*$ is the optimal value for $A$; \emph{k} is the number of iterations. 
\end{proof}

\vspace{-0.3cm}

Theorem \ref{thm:lip} gives us a simple way to set the step size during the
generalized gradient updates and also guarantees the fast convergence rate.

\textbf{Optimization of $\Omega \backslash A = \{ C, R, Q,\pi_1, V_1 \} $}. Each of these parameters is estimated similarly to the approach in \cite{ghahramani1996parameter} by taking the corresponding derivative of the eq.(\ref{eq:Q_func}), setting it to zero, and by solving it analytically. Update rules for $\Omega \backslash A = \{ C, R, Q,\pi_1, V_1 \} $ are shown in Algorithm \ref{alg:mstep}.

The M-Step for optimizing $\Omega$ is summarized in Algorithm \ref{alg:mstep} in Appendix.

\vspace{-0.3cm}

\section{Experiments}
\label{sec:experiments}

\vspace{-0.3cm}

We test our approach on time series data obtained from electronic health records of 4,486 post-surgical cardiac patients stored in PCP database \cite{batal2011pattern,hauskrecht2012outlier,MilosHauskrecht2010,Valko2010}. To test the performance of our prediction model, we randomly select 600 patients that have at least 10 \emph{Complete Blood Count} (CBC) tests \footnote{CBC panel is used as a broad screening test to check for such disorders as anemia, infection, and other diseases.} ordered during their hospitalizations. The three tests used in this experiment are Mean Corpuscular Hemoglobin Concentration (MCHC), Mean Corpuscular Hemoglobin (MCH) and Mean Corpuscular Volume (MCV). These time series data are noisy, their signals fluctuate in time, and observations are obtained with varied time-interval period. 
\nocite{batal2012mining}

%Figure \ref{fig:multiTsVisual} illustrates such time series for one of the patients. The X-axis represents the time indices aligned by hour and the Y-axis shows their normalized values/observations. The solid dots denote the true irregular sampled observations. 

In order to get regularly sampled multivariate time series dataset, we apply an 8-hour discretization on our original multivariate time series dataset and use linear interpolation to fill the missing gaps from discretization. We compare our sparse LDS (SLDS) with ordinary LDS (OLDS) on the above multivariate time series dataset.

To evaluate the performance of our SLDS approach we split our time series for 600 patients into the training and testing sets, such that 50/100 times series form the training data, and 500 are used for testing. 

{\bf Evaluation Metric.} Our objective is to test the predictive performance of our approach by its ability to predict the future value of an observation for a patient for some future time {\em t} given a sequence of patient's past observations. We judge the quality of the prediction using the Average Mean Absolute Error (AMAE) on multiple test data predictions. More specifically, the AMAE is defined as follows: 

\vspace{-0.4cm}

\begin{equation}
\label{eq:mae}
AMAE =  m^{-1} n^{-1} \sum_{i=1}^m \sum_{j=1}^n |y_{i,j} - \hat{y}_{i,j}| 
\end{equation}

\vspace{-0.4cm}

\noindent where $y_{i,j}$ is the \emph{j}th true observation from time series \emph{i}, $\hat{y}_{i,j}$ is the corresponding predicted value of $y_{i,j}$. \emph{m} is the number of time series and \emph{n} is the length of each time series.

\begin{figure}[!htb]
\minipage{0.48\textwidth}
  \includegraphics[width=\linewidth]{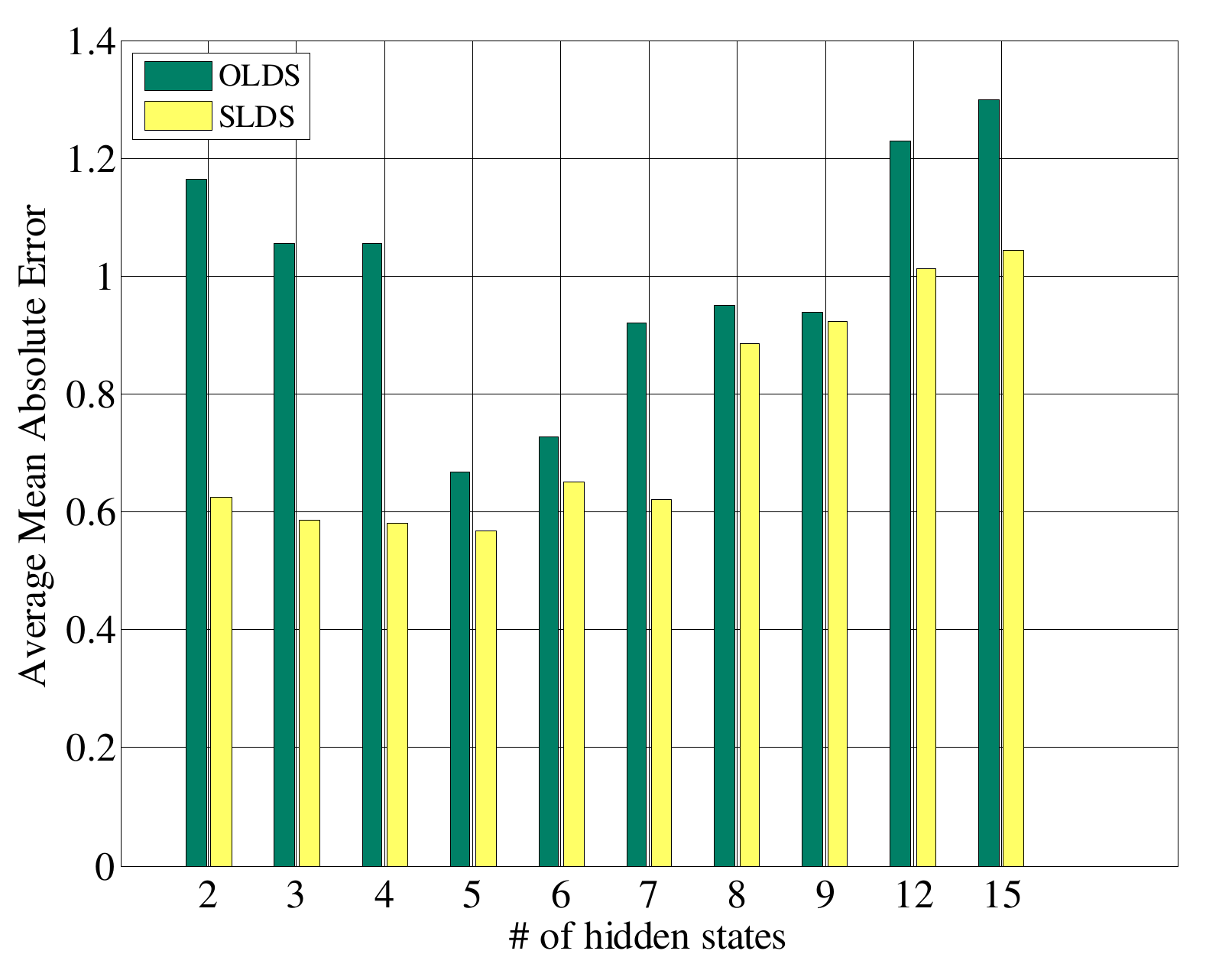}
  \vspace{-0.8cm}
  \caption{AMAE on 50 training examples.}\label{fig:results_50}
\endminipage\hfill
\minipage{0.48\textwidth}%
  \includegraphics[width=\linewidth]{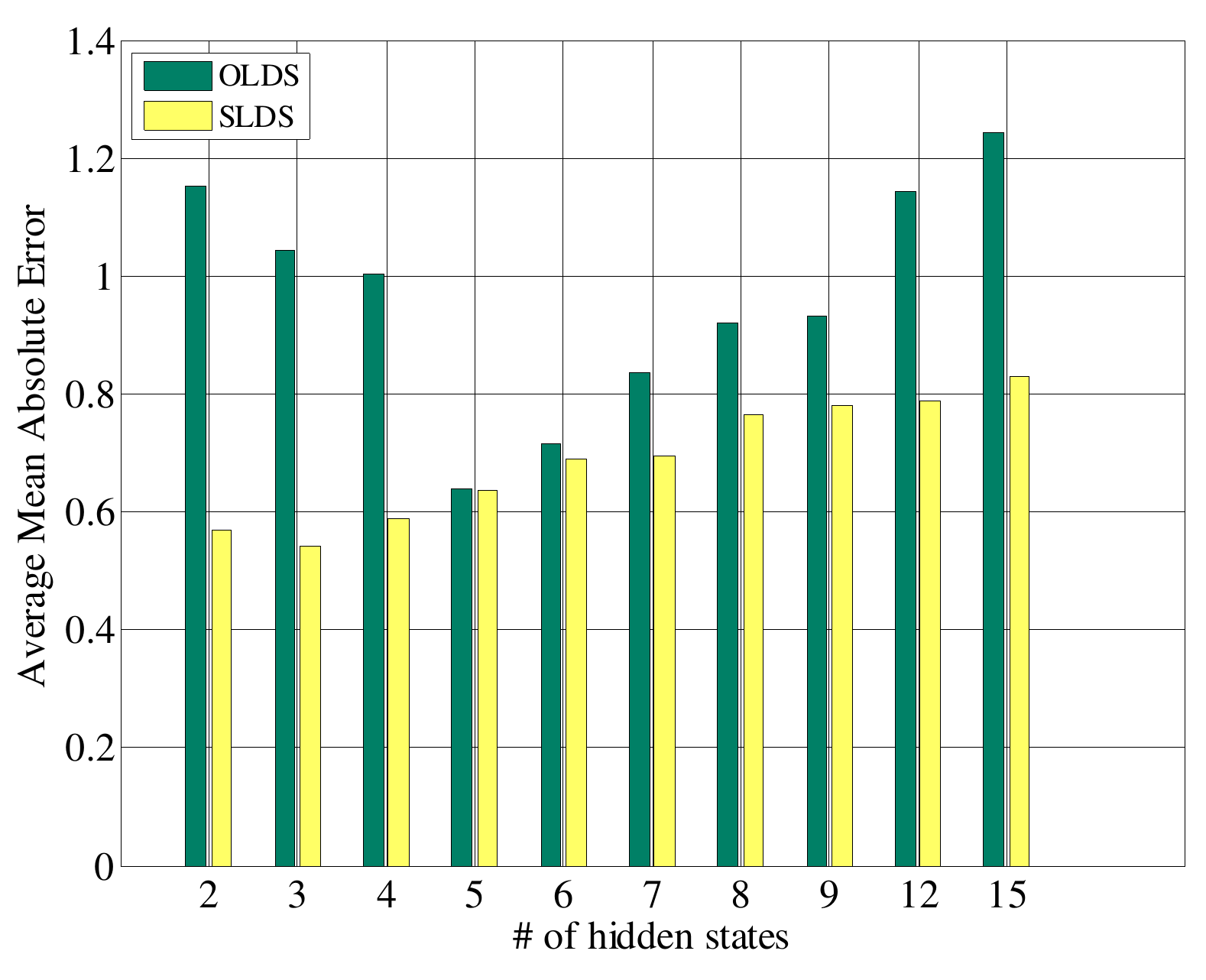}
  \vspace{-0.8cm}
  \caption{AMAE on 100 training examples.}\label{fig:results_100}
\endminipage
\end{figure}

To conduct the evaluation, we use the test dataset to generate various prediction tasks as follows. For each patient $p$ and complete time series \emph{i} for that patient, we calculate the number of observations $n_i^p$ in that time series \emph{i}. We use $n_i^p$ to generate different pairs of indices $(\psi,\phi)$ for that patient, such that $1 \leq \psi < \phi \leq n_i^p$, where $\psi$ is the index of the last observation assumed to be seen, and $\phi$ is the index of the observation we would like to predict. By adding time stamp reading to each index, the two indices help us define all possible prediction tasks that we can formulate on that time series. For each time series \emph{i} from patient \emph{p}, we proceed by randomly picking 5 different pairs of indices (or 5 different prediction tasks) for the total of 2500 predictions tasks (500 x 5 = 2500). For each method, we repeat this random sampling predictions 10 times and we use the Average Mean Absolute Error (AMAE) on these tasks to judge the quality of test predictions. The prediction results are shown in Table \ref{tab:results}, Figure \ref{fig:results_50} and Figure \ref{fig:results_100}.

From Figure \ref{fig:results_50} and Figure \ref{fig:results_100}, we can see that OLDS achieves its lowest prediction error \emph{AMAE} when the number of hidden states is 5. By varying the number of states, the errors for OLDS first improve (till the optimal number of states is reached) and then increase when the number of hidden states exceeds the optimal point. This clearly shows the overfitting problem. The SLDS performs similarly; its performance first impoves and after that it deteriorates. However, its errors deteriorate at slower pace which shows it is more robust to the overfitting problem. Comparing the two methods, the SLDS always outperfroms the OLDS, indicating that the additional sparsity term included in optimization helps it to better fit the underlying structure of the transition matrix.   

\begin{table}[!htbp]
\setlength{\tabcolsep}{3.5pt}
    \caption{Average mean absolute error for OLDS and SLDS with different hidden states sizes.} \vspace{-0.2cm}
\label{tab:results}
\begin{center}
\begin{tabular}{l|ccccccccccc}
\hline
\# of states &  2 & 3 & 4 & 5 & 6 & 7 & 8 & 9 & 12 & 15\\ 
\hline
OLDS(50) & 1.1653 & 1.0553 & 1.0561 & 0.6667 & 0.7268 & 0.9209 & 0.9502 & 0.9402 & 1.2291 & 1.2993 \\
SLDS(50) & 0.6256 & 0.5858 & 0.5821 & 0.5684 & 0.6506 & 0.6200 & 0.8854 & 0.9236 & 1.0134 & 1.0430 \\
\hline
\hline
OLDS(100) & 1.1527 & 1.0427 & 1.0039 & 0.6406 & 0.7153 & 0.8364 & 0.9210 & 0.9327 & 1.1427 & 1.2427 \\
SLDS(100) & 0.5709 & 0.5429 & 0.5889 & 0.6379 & 0.6897 & 0.6949 & 0.7643 & 0.7811 & 0.7874 & 0.8309 \\
\hline
\end{tabular}
\end{center}
\end{table}

\vspace{-0.6cm}

\section{Conclusion}
\label{sec:conclusion}

\vspace{-0.3cm}

In this paper, we have presented a sparse linear dynamical system (SLDS) for multivariate time series predictions. Comparing with the traditional linear state-space systems, SLDS model tries to (1) prevent the overfitting problem and (2) represent additional structure in the transition matrix. Experimental results on real world clinical data from electronic health records systems demonstrated that this novel model achieves errors that is statistically significantly lower than errors of ordinary linear dynamical system. We would like to note that the results presented in this work are preliminary and include only three time series. Further investigation of more complex higher dimensional time-series data is needed and will be conducted in the future. In addition, we would like to study group lasso regularization techniques which we believe would be able to better control the dimensionality of the hidden state space. Finally, we plan to study extensions of our model to switching-state and controlled dynamical systems \cite{hauskrecht1998modeling,kveton2006solving}.

\textbf{Acknowledgement:} This research work was supported by grants R01LM010019 and R01GM088224 from the National Institutes of Health. Its content is solely the responsibility of the authors and does not necessarily represent the official views of the NIH. We would like to thank Eric Heim and Mahdi Pakdaman for useful discussions and comments on this work.

\vspace{-0.4cm}

\bibliographystyle{plain}
{\footnotesize
\bibliography{nips2013}}

\vspace{-0.4cm}

\section*{Appendix} 

\vspace{-0.4cm}

\begin{algorithm}[!phtb]
    \small \caption{EM: M-step for the ($k+1$)th iteration. (We omit the explicit superscript $(k+1)$ for notational brevity.)}
\label{alg:mstep} 
{INPUT:}
\begin{compactitem}
\item Observation sequence $y_t$s, $t = 1,\ldots,T$.
\item Sufficient statistics $\hat{z}_{t|T}$, $M_{t|T}$, $M_{t,t-1|T}$, $i = 1,\ldots,T$ from the ($k+1$)th iteration in E-Step.
\end{compactitem}
{PROCEDURE:}
\begin{algorithmic}[1]
    \STATE Update $\Omega \backslash A$: $C = ( \sum_{t=1}^T y_t \hat{z}_{t|T}^{'} ) ( \sum_{t=1}^T M_{t|T} )^{-1}$, \hspace{0.1cm} $Q = \frac{1}{T-1} ( \sum_{t=2}^T M_{t|T} - A \sum_{t=2}^T M_{t-1,t|T} )$, \hspace{0.1cm} $R = \frac{1}{T} \sum_{t=1}^T ( y_ty_t^{'} - C \hat{z}_{t|T} y_t^{'}  )$, \hspace{0.1cm} $\pi_1 = \hat{z}_{1|T}$, \hspace{0.1cm} $V_1 = M_{1|T} - \hat{z}_{1|T} \hat{z}_{1|T}^{'}$.
    \STATE Initialize $A$, $A = (\sum_{t=2}^T M_{t,t-1|T})(\sum_{t=2}^T M_{t-1|T})^{-1}$.
    \REPEAT
    \STATE Compute the fixed step size $\alpha$, $\alpha = 1/(||{Q}^{-1}||_F\cdot ||\sum_{t=2}^T M_{t-1|T}||_F)$.
    \STATE Compute gradient of $g(A)$, $\bigtriangledown g(A) = {Q}^{-1} A \sum_{t=2}^T M_{t-1|T} - {Q}^{-1} \sum_{t=2}^T M_{t,t-1|T}$.
    \STATE Update $A$, $A = S_{\beta \alpha}(A - t\bigtriangledown g(A) )$.
   \UNTIL{Convergence} 
\end{algorithmic}
{OUTPUT:}
$\Omega^{(k+1)} = \{ A, C, Q, R, \boldsymbol{\pi}_1, V_1 \}$.
\renewcommand*\arraystretch{1.0}
\end{algorithm}

\end{document}